\documentclass[9pt,twocolumn,twoside]{pnas-new}

\articletype{CLASSIFICATION}

\templatetype{pnasresearcharticle} 

\usepackage{hyperref}

\usepackage{booktabs, tabularx, multirow, enumitem}
\usepackage{caption}
\usepackage{subcaption}

\usepackage{url}
\usepackage{xcolor}

\begin{document}

\title{Transfer Learning for Socioeconomic Estimation in Forced-Displacement Settings}

\author[a]{Steven Ndung'u}
\author[b]{Adel Daoud}
\author[c]{Ismael Yacoubou Djima}
\author[c]{Hai-Anh H. Dang}
\author[a]{Patrick Michael Brock}

\affil[a]{The United Nations High Commissioner for Refugees, Denmark}
\affil[b]{Chalmers University, Sweden}
\affil[c]{The World Bank, USA}

\leadauthor{Ndung'u}

\significancestatement{Socioeconomic insight into household vulnerabilities and coping capacities is essential for durable solutions, yet infrequent surveys can quickly become outdated in volatile displacement settings. This study demonstrates how satellite imagery and transfer learning can extend household surveys in displacement contexts by generating spatially explicit, model-based socioeconomic estimates between survey rounds to inform operational prioritization, assessment planning, and targeted field verification across areas inhabited by refugees, internally displaced persons (IDPs), and host communities.}

\authorcontributions{Author's contributions: S.N. conceptualized the study; performed research, analyzed data,  wrote the paper;  P.M.B. conceptualized the study, performed research, wrote the paper; A.D. conceptualized the study, performed research, wrote the paper; I.Y.D. conceptualized the study, performed research, wrote the paper; H.H.D. conceptualized the study,  wrote the paper.}
\authordeclaration{The authors declare no competing interests.}
\correspondingauthor{\textsuperscript{2}To whom correspondence should be addressed. E-mail: ndungust@unhcr.org}


\begin{abstract}

\noindent Progress in inclusive household surveys has strengthened socioeconomic evidence for forcibly displaced populations, providing indispensable benchmarks on living conditions and welfare. However, these surveys remain resource-intensive and periodic, while conditions can change between rounds, particularly in settings affected by fragility, conflict, and violence. More frequently updated, spatially granular complementary evidence is therefore needed to identify where socioeconomic conditions may be changing between survey rounds and to inform operational prioritization. Earth observation and machine learning (EO-ML) offer a scalable source of spatially explicit socioeconomic information. However, tools developed for general populations have not been systematically adapted and evaluated in forced displacement settings, where living conditions, settlement patterns, and displacement impacts may differ substantially. We address this gap by adapting a multimodal spatiotemporal vision transformer, pretrained on Demographic and Health Survey data from approximately 1.2 million households across 36 African countries, to forced displacement and host community settings in South Sudan, Cameroon, and Zambia. We develop and evaluate the updated, adapted model using socioeconomic indices derived from UNHCR Forced Displacement Survey (FDS) and Results Monitoring Survey (RMS) data. Our results show that satellite-derived geospatial covariates explain up to 66\% of the variation in socioeconomic outcomes in camp-intersecting grids, with a mean absolute error (MAE) of 4.37 index points, and 41\% in non-camp-intersecting areas, with an MAE of 5.41. The framework complements and adds value to periodic household surveys by filling critical spatial and temporal data gaps with regularly updated, model-based socioeconomic estimates. These estimates sustain insight between survey rounds and support timely humanitarian prioritization and field verification.

\end{abstract}


\maketitle
\thispagestyle{firststyle}
\ifthenelse{\boolean{shortarticle}}{\ifthenelse{\boolean{singlecolumn}}{\abscontentformatted}{\abscontent}}{}

\Firstpage


Forced displacement gives rise not only to humanitarian emergencies but also to long-term development challenges \cite{robinson2003risks,christensen2009forced}. Many displaced populations live in protracted situations in low- and middle-income host countries, where low investment limits services to host communities, aid to displaced populations often relies heavily on international organizations, and limited statistical capacity leaves persistent socioeconomic data gaps \cite{christensen2009forced,verme2021impact,unhcr2026globaltrends}. Therefore, accurate and timely measurement of household socioeconomic conditions is essential because it underpins evidence-based humanitarian response, poverty reduction, social protection, self-reliance, and durable-solutions planning. Despite this need, comprehensive and frequently updated welfare data for forcibly displaced persons and their host communities remain scarce and structurally difficult to collect \cite{masaki2023data,baal2017obtaining}.

This measurement challenge is amplified by the volatility of many forced displacement settings, where household welfare can change between data collection rounds in response to new arrivals, returns, conflict spillovers, climate shocks, or shifts in humanitarian assistance. The surveys used to measure these conditions are costly and operationally demanding, often leaving gaps of several years between rounds \cite{baal2017obtaining}. As a result, survey evidence may become outdated before the next data collection \cite{fixler2007timeliness}. \Endparasplit \noindent UNHCR’s Forced Displacement Survey (FDS) program illustrates both the value of comprehensive household surveys and the limits of relying on periodic data collection alone. The FDS provides detailed household-level information on refugees, internally displaced persons, returnees, and host communities and is designed to be implemented more frequently than a decadal census, typically every two to five years\footnote{https://www.unhcr.org/what-we-do/reports-and-publications/data-and-statistics/forced-displacement-survey}.  However, important changes can still occur between surveys. For example, the 2023 South Sudan FDS \cite{unhcr2024fdsSouthSudan} coincided with the onset of the April 2023 Sudan crisis, which triggered new refugee arrivals in South Sudan. This limitation is consequential for effective humanitarian and development responses because forcibly displaced populations often experience asset loss, livelihood disruption, restricted mobility, and limited access to services, labor markets, and education \cite{pape2023measuring,alfahal2025conflict,ruiz2015labor}. Complementary approaches to monitoring socioeconomic conditions between survey rounds are therefore essential for timely prioritization, resource allocation, and monitoring of interventions across camps, urban and mixed settlements, and host communities.


Over the past decade, substantial progress has been made in estimating human development indicators using Earth observation and machine learning (EO-ML) \cite{burke2021using,aiken2022machine, chi2022microestimates}. Foundational studies have demonstrated that satellite imagery, combined with deep learning, can predict indicators such as poverty and asset wealth at subnational scales across sub-Saharan Africa \cite{jean2016combining, yeh2020using, zheng2025dynamic, zheng2026satellite, sherman2026global}. By learning the visual signatures of the built environment, infrastructure networks, land use, and land cover from daytime imagery and night-time lights, these models translate pixels (remotely sensed information) into estimates of asset-based wealth, consumption, and poverty \cite{yeh2020using, daoud2023using, adelusi2025utilizing, sherman2026global, zheng2026satellite}. More recently, EO-ML methods have moved beyond static, cross-sectional estimation. By combining multi-decadal satellite time series with recurrent convolutional networks \cite{donahue2015long} and spatiotemporal transformers \cite{dosovitskiy2020image}, they can model changes in socioeconomic indicators over time, including trajectories of poverty reduction, uneven growth, and regional divergence \cite{zheng2025dynamic, pettersson2023timeseries}. These advances have expanded the scope of socioeconomic measurement in settings where census coverage is sparse, surveys are outdated, or operational constraints limit data collection. They enable policymakers not only to target interventions but also to monitor change and assess program impacts at a fine spatial scale.

 Related research has addressed data gaps in forced displacement settings through complementary approaches. Remote-sensing studies have mapped refugee settlements and dwellings, showing that conventional settlement products tend to systematically under-detect such settlements because these areas lack sufficient training and validation data \cite{quinn2018humanitarian,van2021satellite}. Beyond EO-based mapping, machine-learning methods using non-EO data have supported humanitarian-aid targeting \cite{aiken2022machine}, while cross-survey imputation has been used to estimate poverty among refugee populations \cite{sarr2025using}. However, these studies do not evaluate whether socioeconomic representations learned from national household surveys can be transferred and calibrated using displacement-specific survey outcomes. This unresolved question links the broader literature to the three challenges this study addresses.
 
Three interrelated challenges nevertheless remain, each motivating this study. Firstly, a population representation challenge. Existing EO–ML poverty estimation models have been developed and evaluated predominantly using nationally representative household surveys, such as the Demographic and Health Surveys (DHS), or population censuses that primarily represent the general population. Their adaptation to forced displacement settings has received limited attention, particularly at the level of refugee camps and surrounding settlements, where livelihoods and living conditions can vary substantially over short distances. This study therefore extends EO-ML socioeconomic estimation to areas inhabited by refugees and internally displaced persons.

The second is a transferability challenge. Transfer learning reuses representations learned from a source domain to improve prediction in a related target domain, particularly when target-domain data are limited \cite{thrun1998learning,pan2009survey}. Although this approach is increasingly used in EO–ML welfare estimation, existing applications have largely examined transfer across national populations (nondisplaced population) and related geographic contexts \cite{daoud2023using}. Forced displacement settings present a more pronounced domain shift: socioeconomic conditions, settlement morphology, and the relationship between visible infrastructure and household welfare may differ substantially from those observed in general populations. We adapt a model pretrained on national DHS data using displacement-specific FDS and RMS data. We apply a partial fine-tuning procedure; the unresolved empirical question is whether spatial and temporal representations learned from national populations remain informative and can be reliably adapted to forced displacement settings.

The third is an operational use challenge. Even where socioeconomic estimation is technically feasible, it remains unclear whether model-derived estimates can be used to support humanitarian prioritization between survey rounds. During these intervals, humanitarian and development actors must decide where to allocate resources, how to target assistance, and how to monitor progress towards self-reliance and durable solutions, often without a recent and comparable source of household-level evidence. Timely, spatially granular estimates across refugee, internally displaced, returnee, and host community population areas could therefore complement survey data and support decision-making during periods when new survey observations are unavailable.

This paper addresses these three challenges jointly. We adapt the multimodal spatiotemporal vision transformer \cite{pettersson2023timeseries}, pretrained on approximately 1.2 million households from 140 Demographic and Health Surveys conducted across 36 African countries between 1990 and 2020, and apply it to forced displacement contexts in South Sudan, Cameroon, and Zambia. We construct the target socioeconomic index from harmonized Forced Displacement Survey and Results Monitoring Survey data using the standardized principal component analysis methodology developed for displacement settings \cite{greenacre2022principal,leopold2026wealthindex}. This methodology builds on the DHS Wealth Index \cite{rutstein2008dhs} and the International Wealth Index \cite{smits2015international}, while accounting for displacement-specific conditions such as asset loss, dependence on humanitarian assistance, and constraints on land ownership. The Earth-observation inputs comprise Landsat surface-reflectance time series \cite{zhu2019benefits,goward2022landsat}, VIIRS night-time radiance \cite{elvidge2017viirs}, and building features derived from Open Buildings \cite{sirko2023high}. We freeze the early spatial transformer layers while fine-tuning the remaining spatial layers, the temporal transformer, and the regression head using displacement-specific data (FDS/RMS). We train and evaluate the model using a spatially stratified validation protocol across the three country contexts. We then use the resulting model estimates for humanitarian prioritization and socioeconomic monitoring between survey rounds.

This study makes three linked contributions. It extends EO–ML socioeconomic estimation to underrepresented forced displacement populations, tests whether representations learned from national surveys remain useful after displacement-specific adaptation, and evaluates their potential to complement inter-round survey evidence. Building on the multimodal spatiotemporal transformer and DHS pretraining of Pettersson et al. ~\cite{pettersson2023timeseries}, we harmonize FDS and RMS outcomes, integrate them with Landsat, VIIRS, and building data, and evaluate a single cross-country model across South Sudan, Cameroon, and Zambia. Fine-tuning yields country-level test $R^2$ values of 0.62, 0.28, and 0.44, respectively, and 0.66 for camp-intersecting grids.

The paper proceeds as follows. Section~\ref{sec:results} reports the results. Section~\ref{sec:discussion} discusses implications for humanitarian operations, methodological limitations, and directions for future work. Section~\ref{sec:data} describes the survey, satellite, and auxiliary geospatial data used in this study. Section~\ref{sec:methodology} presents the methodology: data harmonization procedures applied across the three country contexts, model architecture, the transfer learning strategy, and performance metrics for model evaluation. Finally, Section~\ref{sec:conclusion} presents conclusions.

\section{Results}
\label{sec:results}

 We first examine the distribution and comparability of the socioeconomic index across survey sources and population groups. Then we evaluate the fine-tuned model's predictive performance across settlement areas, and finally demonstrate its use for generating model-based estimates between survey rounds over time and space.

\subsection{Socioeconomic index}
\label{sec:socioeconomic_index}

\begin{figure*}[p]
    \centering

    \includegraphics[
        width=\textwidth,
        height=0.84\textheight,
        keepaspectratio,
    ]{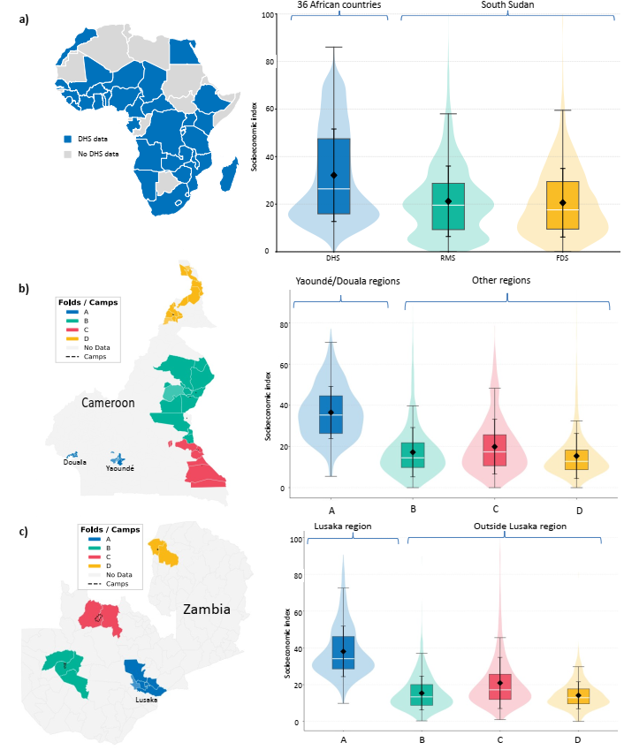}

    \vspace{0.3em}

    \caption{Socioeconomic index distributions across the three study countries. (a) Left: the 36 African countries with DHS data used for pretraining (map). Right: the DHS wealth-index distribution compared with the South Sudan socioeconomic indices constructed from RMS and FDS data. Both displacement distributions are concentrated toward lower index values relative to the DHS reference. (b) Cameroon and (c) Zambia: index distributions across spatially disjoint regional folds A-D, with maps showing the extent of each fold. Camp areas are shown as polygons with dotted boundaries. Note that in a) South Sudan is shown as a distributional comparison rather than a regional map to contextualize the distinct FDS and RMS survey sources relative to the DHS pretraining distribution.}
    
    \label{fig:socioeconomic_distribution_cmr_ssd_zmb}
\end{figure*}

The FDS is the central household survey dataset in this study and provides the primary displacement specific socioeconomic labels for model development and evaluation. It captures detailed information on living conditions, welfare, and asset ownership among forcibly displaced and host community populations. However, the available FDS data remain limited in spatial and temporal coverage. We therefore explored whether RMS data could provide an aligned, complementary source of socioeconomic labels. Although the two survey instruments are largely similar, RMS data do not include asset variables (the remaining variables capturing socioeconomic conditions are similar). Consequently, the FDS index includes detailed information on asset ownership, whereas the RMS index is constructed primarily from housing quality, access to services, and food consumption. As shown in Fig.~\ref{fig:socioeconomic_distribution_cmr_ssd_zmb}a, the two indices exhibit broadly aligned distributional patterns while retaining differences associated with their respective population groups and survey contexts (Table~\ref{tab:combined_distribution}). This alignment supports using RMS data alongside the FDS to increase the spatial and temporal coverage available for fine-tuning.

The DHS reference distribution provides a benchmark for understanding the domain shift between the national survey data used for pretraining and the displacement indices used for fine-tuning. The same panel in Fig.~\ref{fig:socioeconomic_distribution_cmr_ssd_zmb}a shows that the normalized displacement indices are more concentrated towards the lower end of the socioeconomic scale, whereas the DHS reference distribution, representing general populations across 36 African countries, has a higher central tendency and a longer upper tail. Because these indices are derived from different surveys and populations, these patterns should not be interpreted as a direct cardinal comparison of welfare levels. Instead, they illustrate broad distributional differences between the DHS pretraining domain and the displacement-specific fine-tuning domain, thereby motivating the adaptation strategy described in Section~\ref{subsec:transfer_learning}.

Beyond these cross-survey patterns, disaggregating the index by spatial region (Fig.~\ref{fig:socioeconomic_distribution_cmr_ssd_zmb}b and~\ref{fig:socioeconomic_distribution_cmr_ssd_zmb}c) reveals considerable geographic variation. Regions anchored by major urban centers, namely Yaoundé and Douala in Cameroon and Lusaka in Zambia, have substantially higher and wider index distributions than other areas. This pattern may reflect greater access to labor markets, services, and infrastructure among refugee households and host community households in urban areas and their peripheries.

Within this broader regional variation, the three Zambian camp settings exhibit substantially overlapping socioeconomic index distributions: Mayukwayukwa ($n=684$), Meheba ($n=1{,}025$), and Mantapala ($n=450$). Their mean index values are similar at 17.3, 17.6, and 15.9, respectively, and their distributions overlap considerably (Fig.~\ref{fig:socioeconomic_distribution_zambia_camps} and Table~\ref{tab:zambia_camp_socioeconomic_summary} in the Appendix). Despite differences in sample size, location, and history, the three settlements therefore exhibit broadly comparable socioeconomic profiles. Relative to the wider urban and rural variation, they occupy a comparatively low and compressed portion of the index distribution.

\subsection{Model performance}
\label{sec:Model_performance}

We evaluate model performance in three stages. First, we assess the direct transfer of the pretrained DHS model without adaptation. Next, we report the fine-tuned model's performance on the pooled evaluation data and, finally, examine its generalization across the three country contexts.

As a zero-shot baseline, we applied the pretrained DHS model to the displacement data without updating its parameters. The resulting predictions were concentrated within a narrow range near the mean of the source training distribution, producing near-zero or negative $R^2$ values (Table~\ref{tab:zero_shot_model_performance_summary} in the Appendix). The pretrained model, therefore, does not transfer reliably to forced displacement settings without adaptation. This result does not imply that the pretrained backbone contains no useful information. Rather, it indicates that the DHS-calibrated prediction mapping is poorly aligned with the target distribution and requires domain-specific fine-tuning.

\begin{figure}[ht]
    \centering
    \footnotesize
        \includegraphics[
            width=\linewidth]{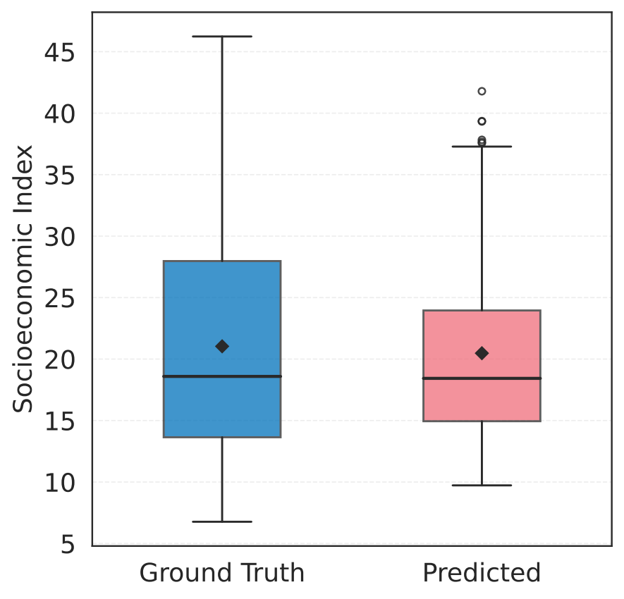}
        \caption{Observed and predicted socioeconomic-index distributions, illustrating prediction shrinkage toward the center.} 
\label{fig:model_shrinking}
\end{figure}

\begin{figure*}
   \centering
    \footnotesize

    \setcounter{subfigure}{2}

    \begin{subfigure}[t]{0.45\linewidth}
        \centering
        \includegraphics[
            width=\linewidth]{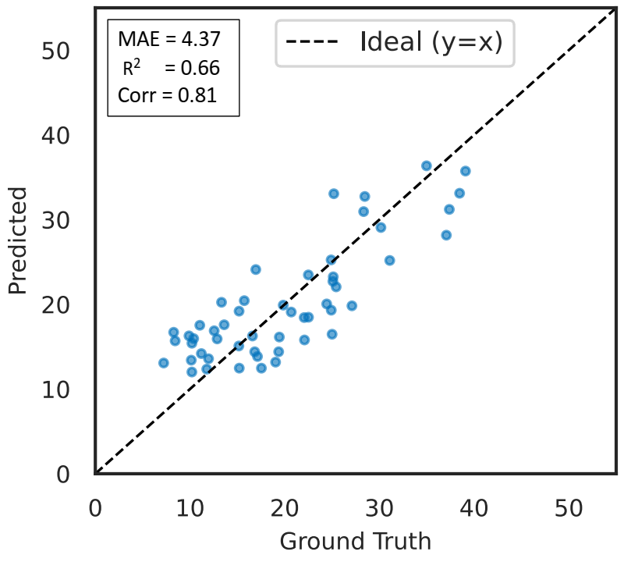}
        \caption{Camp-intersecting grid settlements model performance.}
    \end{subfigure}
    \hfill
    \begin{subfigure}[t]{0.45\linewidth}
        \centering
        \includegraphics[
            width=\linewidth]{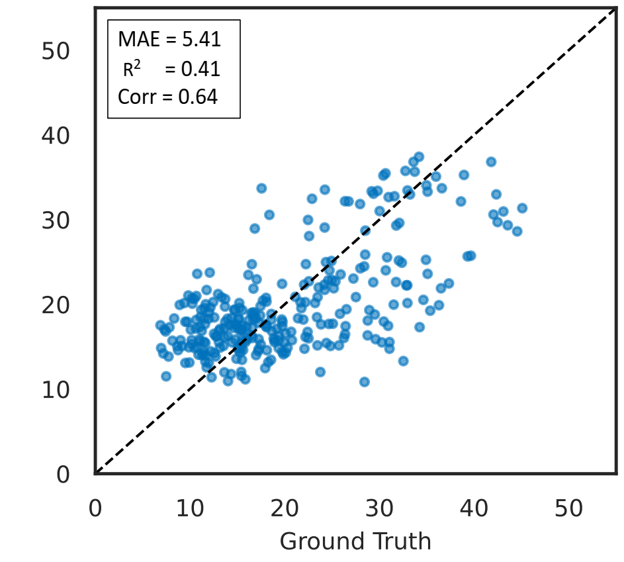}
        \caption{Non-camp-intersecting grid areas model performance.}
    \end{subfigure}

    \caption{\footnotesize
    Predictive performance on the pooled held-out test data by settlement context. Panel (a) shows grids intersecting refugee or IDP camp boundaries, and panel (b) shows non-camp-intersecting grids.}
    \label{fig:valid_test_plots}
\end{figure*}

Following fine-tuning, the best-performing model, selected using the validation set, achieved an $R^2$ of 0.40 and a mean absolute error (MAE) of 5.49 index points across the pooled evaluation sample. The predictions broadly track the observed values but exhibit a compressed range relative to the identity line (as seen in Fig.~\ref{fig:valid_test_plots}a and b). The model tends to hedge in regions of greater uncertainty, overpredicting lower socioeconomic values and underpredicting higher ones, as shown in Fig.~\ref{fig:valid_test_plots}b. This regression-to-the-mean pattern is common in models trained to predict continuous outcomes, including wealth, from satellite imagery, where predictions tend to shrink toward the mean of the training distribution \cite{berglund2012regression,pettersson2025debiasing} ( Fig.~\ref{fig:model_shrinking}). Compared with the narrowly distributed zero-shot predictions, fine-tuning substantially improves both the range and agreement of predictions with the observed values.

Under a spatially stratified validation protocol (as described in Section~\ref{Experimental_setup}), the test grids included displacement-affected populations, comprising refugees living both inside and outside camps, as well as host communities. Across the three country contexts, the model achieved an $R^2$ of 0.62 in South Sudan, 0.44 in Zambia, and 0.28 in Cameroon, with corresponding correlations of 0.89, 0.72, and 0.56, respectively (Table~\ref{tab:model_performance_refugees_groupB}). South Sudan also recorded the lowest MAE at 3.55 index points, compared with 4.53 in Zambia and 5.48 in Cameroon. The larger South Sudan sample, which combines FDS and RMS observations, contributes to its stronger performance. The lower performance in Cameroon and Zambia highlights the greater difficulty of extrapolating to spatially disjoint and heterogeneous regions (out-of-distribution samples). Nevertheless, the positive $R^2$ and correlation values across all three countries indicate that the adapted representation retains useful predictive information beyond the regions used for model training. 

\begin{table}[!h]
\centering
\footnotesize
\caption{Model performance in South Sudan, Cameroon and Zambia on the test data.}
\label{tab:model_performance_refugees_groupB}
\begin{tabular}{p{1.7cm}p{0.8cm} p{0.8cm} p{0.8cm}p{0.8cm}p{1.0cm}}
\toprule
Country & MAE & MSE & RMSE & $R^2$ & Correlation \\
\midrule
South Sudan & 3.55 & 16.36 & 4.04 & 0.62 & 0.89 \\
Cameroon & 5.48 & 44.51 & 6.67 & 0.28 & 0.56 \\
Zambia & 4.53 & 34.07 & 5.84 & 0.44 & 0.72 \\
\bottomrule
\end{tabular}
\end{table}

The central aim of this work is to adapt a model trained on general national populations to forcibly displaced populations. We therefore analyze the results by pooling held-out data from all three countries, excluding training grids, and stratifying the evaluation sample into the two subgroups that define the adaptation problem: grids intersecting refugee and IDP camp boundaries, and grids situated among surrounding host communities. Evaluating these subgroups separately tests whether domain transfer succeeds across distinct forced displacement contexts, where differing socioeconomic and spatial dynamics may affect both model performance and operational utility \cite{roscher2024better}. The subgroup evaluation shows higher predictive performance in camp-intersecting grids than in non-camp-intersecting grids, although differences in outcome variability, sample composition, and household density may partly explain this pattern. In camp-intersecting grids, the model attains an $R^2$ of 0.66 with a mean absolute error of 4.37 index points, against 0.41 and 5.41, respectively, for host community grids (Table~\ref{tab:model_performance_summary}; Fig.~\ref{fig:valid_test_plots}c and ~\ref{fig:valid_test_plots}d). The stronger performance in camp-intersecting grids may reflect more consistent characteristics visible in Earth observation data, including dwelling density, settlement layout, infrastructure, and changes in surrounding vegetation. These features allow the model to learn relationships between settlement morphology and socioeconomic conditions. By contrast, non-camp-intersecting grids span a wider range of socioeconomic conditions and settlement types, resulting in higher residual errors. Nevertheless, the fine-tuned model outperforms the zero-shot baseline and provides meaningful spatial discrimination in both camp-intersecting and non-camp-intersecting areas.

The subgroup evaluation demonstrates that the model captures distinct socioeconomic variation in camp-intersecting settings with accuracy exceeding its aggregate performance, precisely in areas where between-round monitoring gaps are most acute. One qualification applies, however, to interpreting camp-intersecting grid areas. Since each grid cell spans $6.72 \text{ km} \times 6.72 \text{ km}$, considerably larger than the footprint of most camps, grids intersecting camp boundaries also encompass host communities in the immediate vicinity, within a radius of up to $3$–$5 \text{ km}$. Camp-intersecting grid areas therefore represent camp populations together with their proximate hosting environments, rather than households within camps in isolation.

\begin{table}[!h]
\centering
\footnotesize
\caption{Model performance summary.}
\label{tab:model_performance_summary}
\begin{tabular}{p{3.45cm}p{0.4cm} p{0.4cm} p{0.4cm}p{0.4cm}p{0.7cm}}
\toprule
Settlement areas & MAE & MSE & RMSE & $R^2$ & Cor. \\
\midrule
Camp-intersecting grids   & 4.37 & 25.15 & 5.01 & 0.66 & 0.81 \\
Non-camp-intersecting grids & 5.41 & 46.09 & 6.79 & 0.41 & 0.64 \\
\bottomrule
\end{tabular}
\end{table}


       
        


        
        

\subsection{Using the model: hindcasting and nowcasting}
\label{sec:hindcasting_and_nowcasting}

Beyond cross-sectional estimation, the fine-tuned model can serve as a monitoring tool between survey rounds or when follow-up surveys are unavailable, for example, due to lack of funding. Because Earth observation inputs are acquired regularly and are comparable across regions and years, the model can be applied retrospectively to estimate conditions during the most recent completed survey year (hindcasting) and to more recent imagery to produce updated model-based estimates where survey data are unavailable (nowcasting). We illustrate both modes in South Sudan using the 2023 FDS as the ground-truth baseline. These estimates complement survey and operational evidence by indicating medium-term changes in observable features, such as settlement density, cropping patterns, and infrastructure, over annual or semi-annual intervals. 

\begin{figure*}[t]
    \centering
    \footnotesize     \includegraphics[width=\linewidth]{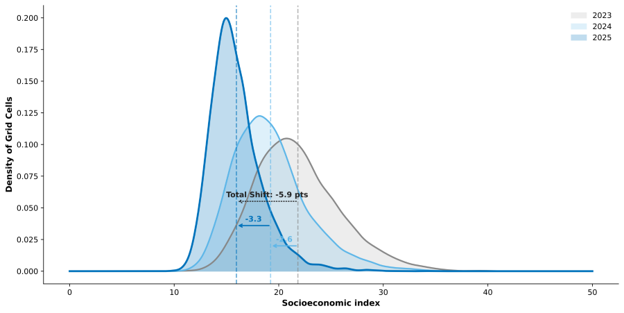} 
    
    \caption{South Sudan socioeconomic index distributions for 2023, 2024, and 2025. Density curves show the model-predicted grid-level index for each year, with the model held fixed and only the Earth observation inputs varied. The distribution shifts progressively toward lower values, with the mean declining by 5.9 index points from 2023 to 2025. Note: These are unvalidated temporal projections and should not be interpreted as observed welfare change.}
    \label{fig:hindcast_nowcast}
\end{figure*}

Holding the fine-tuned model fixed and varying only the Earth observation inputs, we run it on imagery from 2023, 2024, and 2025. The estimates span all grids in the study area where the FDS and RMS data were collected (country administrative level 3), covering refugees living inside formal camps, refugees residing outside camps, and surrounding host communities, so the resulting trajectory reflects the full displacement-affected study area rather than any single group. The hindcast produces model-based estimates for the 2023 baseline, and applying the model to later imagery produces a predicted downward shift in the socioeconomic index through 2024 and 2025 (Fig.~\ref{fig:hindcast_nowcast}), with the mean socioeconomic index declining by 5.9 points relative to the baseline. The direction and geography of this shift are consistent with the documented consequences of the war in Sudan and its cross-border spillover.

These projections illustrate the approach's operational potential by identifying areas of possible socioeconomic change for field verification prior to any follow-up survey rounds.

\section{Discussion}
\label{sec:discussion}

This study provides, to our knowledge, the first systematic evaluation of transfer learning from national nondisplaced populations to forcibly displaced populations, addressing the three challenges identified at the outset, that is, population representation, transferability, and operational use. The central finding is that spatial and temporal representations learned from national populations contain relevant information for refugee, internally displaced, and host community settings but require displacement-specific adaptation. The zero-shot model produces narrowly distributed predictions with near-zero or negative ($R^2$) values, whereas after fine-tuning with FDS and RMS data, satellite-derived geospatial covariates explain up to 66\% of the variation in socioeconomic outcomes across camp-intersecting grids. This improvement provides the main evidence that useful learning transfers despite the domain shift. Annual model-based updates also indicate medium-term socioeconomic changes over time and space that are consistent with documented ground realities. This performance in both camp-intersecting and non-camp-intersecting areas, along with its alignment with operational evidence, supports further evaluation of the model for flagging areas of concern, supporting area-based prioritization, and guiding field verification between survey rounds.

In other comparative studies, EO-ML models have achieved strong but context-dependent performance in predicting asset wealth from satellite imagery. Yeh et al. ~\citep{yeh2020using} report that models trained on publicly available multispectral imagery achieve approximately $R^2=0.70$ of the variation in village-level asset wealth in held-out African country-years. Zheng et al. ~\citep{zheng2025dynamic} report country-level transformer performance of $R^2=0.83$ in Malawi, $R^2=0.70$ in Mozambique, and $R^2=0.62$ in Madagascar using census-derived asset wealth labels; in Burkina Faso, where the effective sample size is smaller, performance reaches $R^2=0.63$ with XGBoost \cite{chen2016xgboost} and $R^2=0.57$ with a satellite-only transformer. For wealth-change prediction, the same study reports temporal performance of $R^2=0.52$ for decadal change in Malawi and $R^2=0.42$ in Mozambique. Similarly, Pettersson et al. ~\citep{pettersson2023timeseries} show that explicitly modeling satellite image time series improves generalization, explaining $R^2=0.72$ of wealth variation across held-out countries and $R^2=0.75$ across held-out time spans. Against these benchmarks, our performance of $R^2=0.66$ for spatial variation in survey-measured welfare within camp-intersecting grids is broadly comparable, despite the distinct and data-constrained nature of forced displacement settings. These benchmarks, however, typically rely on substantially larger training samples from nationally representative household surveys and are evaluated in populations similar to those used for model development. By contrast, our application targets a smaller, policy-critical domain that is largely absent from national sampling frames, structurally different from DHS/national populations, and characterized by distinctive settlement forms, humanitarian assistance regimes, and constrained livelihood systems. Although differences in targets and evaluation designs limit direct comparison, camp-intersecting performance falls within the broad range reported in related EO-ML wealth-estimation studies.

Small-area estimation, cross-survey imputation and phone-based prediction provide relevant operational alternatives to EO-ML. Elbers et al. \cite{elbers2003micro} combined household surveys with census covariates to estimate local poverty, while related cross-survey approaches used shared predictors to estimate welfare in datasets without income or consumption measures, including UNHCR administrative records for refugees \cite{sarr2025using}. Phone-based approaches instead link survey responses to mobile phone metadata to predict wealth \cite{smythe2022geographic}. Our framework complements these methods by using spatially extensive EO inputs to estimate socioeconomic conditions beyond surveyed locations, without requiring household administrative or phone records at each prediction location. Its contribution is therefore to extend geographic information coverage, while remaining dependent on survey calibration.

Operationally, the relative value of these approaches depends on available data, update requirements, and the intended decision. Reusing open-access satellite imagery may enable inexpensive updates after model development, while imputation can similarly draw on existing surveys and administrative records. Blumenstock et al. \cite{blumenstock2015predicting}, for example, report a calibration survey costing approximately US \$12,000 and taking four weeks. However, our study does not establish comparative costs or delivery times. The findings support further evaluation for area-level screening and prioritizing field assessments, while temporal accuracy and targeting benefits remain to be established.

Forced displacement surveys are rich in content but limited in spatial (by definition, given the heterogeneous distribution of displaced settlements) and temporal coverage (given that inclusion in national surveys is only just gaining momentum), and the number of grid-level training observations remains small relative to the scale of modern deep learning models. In this context, the pretrained DHS model provides a broad socioeconomic representation learned from many countries and survey years, while partial fine-tuning adapts this representation to the specific morphology and welfare distribution of forced displacement and host community settings. This design follows the logic emerging in recent work on large-scale pretrained models for wealth mapping: broad pretraining can reduce dependence on expensive labeled data, while task-specific adaptation remains necessary to handle local domain shifts.

The model is intended to complement, not replace, household surveys. As new satellite imagery becomes available, it can generate estimates at predefined intervals, providing regularly updated, model-based indications of where socioeconomic conditions may be changing between survey rounds. The South Sudan estimates (Fig.~\ref{fig:hindcast_nowcast}), for example, should be interpreted as model-based projections that are consistent with documented crisis dynamics, rather than as direct measurements of welfare change. Validation against future FDS and RMS rounds, together with data from South Sudan’s National Household Budget Survey, will be essential for determining whether predicted changes correspond to measured changes on the ground. We will also assess whether the modeled spatial and temporal patterns are consistent with related indicators, such as food-security measures from the Integrated Food Security Phase Classification (IPC)\footnote{\url{https://www.ipcinfo.org/}}, while recognizing that these capture a distinct but related dimension of socioeconomic well-being. Operationally, these outputs are best suited to identifying areas that may be experiencing deterioration and prioritizing them for further assessment and field verification, rather than serving as official poverty or welfare estimates.

Future work can build on these findings as new data accumulate. A key limitation of the current evaluation is that it is fully blocked in space but only partially in time. We keep Admin-3 regions separate across the training, validation, and test sets, ensuring the model is evaluated in geographically unseen areas. Although the combined FDS and RMS data introduce observations from different survey years, they do not provide repeated measurements at the same locations across multiple rounds. Therefore, we cannot hold out a later survey wave and test whether predicted changes correspond to observed socioeconomic changes over time. This limits the temporal validation of the hindcasting and nowcasting applications. Additional FDS and RMS rounds will enable fully blocked spatiotemporal evaluation using later survey waves as independent test data. On the imagery side, Harmonized Landsat and Sentinel-2 (HLS) products could increase observation frequency relative to Landsat’s 16-day revisit cycle, improving sensitivity to seasonal and shorter-term physical changes in vegetation, settlement expansion, flooding, and infrastructure development \cite{ju2025harmonized}. Higher-resolution imagery may further support finer-scale estimation in dense camp and urban settlement contexts.

\section{Data}
\label{sec:data}
This study integrates household survey data from three forced displacement contexts in sub-Saharan Africa with Earth observation data to construct a harmonized dataset that links socioeconomic outcomes to remotely sensed characteristics of settlements and their environs. The following subsections describe the survey-derived reference data, satellite imagery, and building-derived features used in the study.

\subsection{Survey reference data}
\label{subsec:survey_reference_data}

The primary survey reference data for this study come from UNHCR’s flagship Forced Displacement Survey, which has been implemented in four country contexts to date: South Sudan in 2023, Pakistan in 2024, Cameroon in 2024, and Zambia in 2025\footnote{https://www.unhcr.org/media/forced-displacement-survey-zambia-2025}. The FDS is a multi-topic household survey that collects detailed information on living standards, socioeconomic well-being, access to services, and vulnerability among refugees, former refugees, asylum seekers, and host communities. Its standardized design enables comparability across countries and over time, making it a suitable empirical basis for constructing harmonized welfare indicators in forced displacement settings.

This work focuses on the three African FDS country contexts: South Sudan, Cameroon, and Zambia. In South Sudan, we further combine FDS data with the Results Monitoring Survey (RMS), as summarized in Table~\ref{tab:combined_distribution}. Similar to the FDS, the RMS collects household-level information on living conditions, well-being, and rights. In the South Sudan context, it focuses primarily on internally displaced persons, thereby extending the coverage of displacement-affected populations beyond the FDS sample. Combining these sources increases the spatial, temporal, and demographic coverage of the analysis and supports the development of a more comprehensive ground-truth dataset for model training and evaluation.

\begin{table}[ht]
\centering
\footnotesize
\caption{\footnotesize Survey reference-data distribution for South Sudan, Cameroon, and Zambia. FDS denotes the Forced Displacement Survey, and RMS denotes the Results Monitoring Survey. In the dataset column, the year of data collection is shown in parentheses.}

\begin{tabular}{p{1.5cm}p{1.3cm} p{2.5cm} p{0.5cm}p{0.5cm}}
\toprule
 Country & Dataset & Group & Count & \%  \\
\midrule
\multirow{4}{*}{South Sudan}  & \multirow{2}{*}{FDS (2023)} & Host community & 992 & 32.23  \\
 & & Refugees & 2086 & 67.77 \\
 \cmidrule{2-5}
& \multirow{2}{*}{RMS (2024)} & IDPs & 3974 & 63.44 \\
 & & Refugee returnees & 2290 & 36.56 \\
 \midrule
\multirow{2}{*}{Cameroon}  & \multirow{2}{*}{FDS (2024)} & Host community   & 1578 & 34.70 \\       
 & & Refugees  & 2969 & 65.30 \\
  \midrule
\multirow{3}{*}{Zambia}  & \multirow{3}{*}{FDS (2025)} & Host community   & 1194 & 29.55 \\       
 & & Refugees  & 2131 & 52.73 \\
 & & Former Refugees  & 716 & 17.72 \\
\bottomrule
\end{tabular}
                  
\label{tab:combined_distribution}
\end{table}

\subsection{Satellite imagery}
\label{subsec:satellite_imagery}

We use satellite-derived covariates to provide a spatially consistent representation of the physical and economic environments surrounding surveyed households. Two complementary Earth observation data streams are used: daytime multispectral optical imagery from Landsat \cite{zhu2019benefits,goward2022landsat} and night-time light observations from the Visible Infrared Imaging Radiometer Suite (VIIRS) \cite{elvidge2017viirs}.

For the daytime optical component, we use Landsat Collection 2 Level-2 surface reflectance imagery. These products are atmospherically corrected and provide calibrated measurements that are comparable across scenes and acquisition dates. We filter the images using quality assessment information to remove clouds, cloud shadows, and other contaminated pixels. We retain the visible, near-infrared, and shortwave-infrared bands, which are suitable for distinguishing vegetation, bare soil, surface water, and broad built-environment characteristics relevant to settlement morphology. In forced displacement settings, these features are informative because welfare-related differences may be reflected in settlement layout, housing density, vegetation condition, proximity to services, and the surrounding livelihood environment.

We incorporate night-time light data from the VIIRS Day/Night Band to complement the daytime optical imagery. We use night-time radiance as a proxy for artificial lighting and economic activity, rather than as a direct measure of welfare. It provides contextual information on electricity access, settlement intensity, infrastructure concentration, and proximity to active service or commercial areas.

Each survey location is represented by a fixed spatial grid of $6.72 \text{ km} \times 6.72 \text{ km}$. This corresponds to a $224 \times 224$-pixel Landsat image patch at 30 m spatial resolution. VIIRS data are provided at 15 arc-second spatial resolution and spatially aligned to the same geographic extent. This alignment ensures consistency with the input structure of the pretrained model described in Section~\ref{subsec:model_architecture} and provides sufficient spatial context to capture the settlement core and its surrounding environment, including nearby agricultural land, roads, water sources, vegetation gradients, and other contextual features that may not be directly observable at the household location.

For the three countries, Landsat and VIIRS observations are drawn from the same calendar year as the corresponding survey, ensuring temporal alignment between the Earth observation inputs and the survey-derived reference outcomes.

\subsection{Building features and settlement morphology}
\label{subsec: building_footprints_and_settlement_morphology}

To enrich the representation of settlement structure in forced displacement contexts, we incorporate building-level features derived from the Open Buildings 2.5D Temporal dataset \cite{sirko2023high}. This dataset provides annual rasters of building presence, building height, and fractional building count derived from Sentinel-2 imagery, at an effective resolution of 4 m. We use the 2023 layer. 

We extract three complementary bands: building presence, building height, and building fractional count. These variables describe the spatial extent of the built environment, the vertical structure of buildings, and the estimated building density within each pixel. They directly represent settlement morphology, complementing the broader land-surface information captured by Landsat imagery and the activity-related signal from VIIRS night-time lights.

The original pretrained model did not include building features, as it was trained on satellite imagery aligned with Demographic and Health Survey data. However, we include them in this study as an additional displacement-relevant covariate because the target survey data are limited in both sample size and temporal depth. 

\subsection{Integrating satellite data with survey data}
\label{subsec:eo_survey_data_integration}

We integrate survey and satellite data using a common spatial grid. Each household survey location is assigned to a fixed $6.72 \text{ km} \times 6.72 \text{ km}$ grid cell, from which the corresponding Landsat optical imagery, VIIRS night-time lights, and building features are extracted. Household-level socioeconomic indicators are then aggregated to the grid level, such that each grid cell represents a single observation for model training and evaluation. This is achieved by averaging the household-level label values for all geopoint locations overlapping each grid tile. We also impose a minimum threshold of five households per grid to reduce label noise and exclude grid averages that we consider outliers (for example, those with scores of less than 5). The grid-based integration generates 211 observations in South Sudan, 238 in Cameroon, and 159 in Zambia. Notably, we adopt a grid-based approach for two reasons: to preserve household privacy and to ensure alignment with the model input structure, as described in Section~\ref{subsec:model_architecture}.

The distribution of survey samples across grid cells is uneven. Some grids around camp settlements contain more than 50 household observations, whereas others, particularly those covering host community areas, contain only a few households. To increase variation and better represent high-density refugee settlements, we apply a sliding-window strategy. This procedure involves systematically shifting the extraction frame around a central grid covering a densely populated area (illustrated as the solid boundary - red in Fig.~\ref{fig:FDS_data_aug} in the Appendix). By applying horizontal and vertical shifts, we generate spatially offset snapshots, represented by the dotted (yellow) and dash-dotted (blue) grids, that retain significant overlap with the original view. By augmenting the dataset with additional spatial variations, this approach increased the total count to 1{,}123 grids. Because we adopted a spatially stratified validation strategy (see Section~\ref{Experimental_setup} for details), we assigned regional partitions before generating shifted windows, and we retained all windows derived from the same original grid cell within the same training, validation, or test partition to prevent spatial leakage. Each shifted window captures a slightly different spatial context around the same settlement, providing additional spatially perturbed examples for model training.

\section{Methodology}
\label{sec:methodology}

This section describes the methodological framework used to construct the socioeconomic index, integrate survey and Earth observation data, design the spatial validation strategy, and adapt the pretrained spatiotemporal transformer to forced displacement settings through partial fine-tuning.

\subsection{Construction of the socioeconomic index}
\label{subsec:socioeconomic_index_construction}

Household-level welfare indices provide a tractable measure of economic status where monetary poverty data are unavailable, and are particularly valuable in forced displacement settings where national surveys rarely include refugees \cite{leopold2026wealthindex}. We construct the socioeconomic index following the standardized methodology developed for Forced Displacement Survey data \cite{leopold2026wealthindex}, which builds on the DHS Wealth Index \cite{rutstein2008dhs, rutstein2015steps} and the International Wealth Index \cite{smits2015international}, with adaptations for the structural features of displacement contexts such as asset loss, dependence on humanitarian assistance, and constraints on land ownership.

The index is derived from household-level indicators capturing multiple dimensions of welfare: asset ownership, including items such as television, radio, and car; housing quality, classified according to wall, roof, and floor materials; access to basic services, including water, electricity, and sanitation; food security, captured through household consumption indicators; and land-related variables such as ownership status and plot size.

For South Sudan, we complement the FDS data with the RMS, which shares substantial structural similarities with FDS in question wording and response coding, and provides an additional source of survey data for model development and validation. The RMS does not, however, cover asset ownership variables, which constrains the set of variables available for direct harmonization. Because household welfare is multidimensional and extends beyond income and asset ownership, and because housing quality and access to basic services are established proxies for material living standards in data-scarce settings \cite{rutstein2008dhs, smits2015international}, we construct a parallel index for the RMS observations using only the variables common to and aligned with the FDS instrument, with matching question semantics and value labels. In practice, this parallel construction draws on housing quality, access to basic services, and food consumption variables, which both instruments include with harmonized coding. This preserves comparability across the pooled sample while serving as a proxy for the dimensions not directly covered. We refer to the resulting measure as a socioeconomic index rather than an asset index, since its construction extends beyond asset variables alone.

Prior to running the principal component analysis, we apply a series of preprocessing steps to reduce heaping, in which many households receive identical scores due to limited variation in the input variables. Specifically, we merge rare and uncommon response categories, drop variables with little or no variance, transform skewed continuous variables into categorical or binary form, and engineer additional discriminatory variables such as a living conditions score and an asset count score. We then apply principal component analysis to extract the main latent socioeconomic dimension and linearly rescale the first principal component to a 0-100 index for interpretability, with lower scores indicating poorer households and higher scores indicating relatively better-off households. We refer the reader to \cite{leopold2026wealthindex} for further details on the index construction and on detailed analyses of the South Sudan socioeconomic index.

\subsection{Experimental setup}
\label{Experimental_setup}

The experimental design uses a four-fold spatially blocked validation strategy to evaluate model generalization across geographically distinct forced displacement settings. This reflects a data-centric approach in which the structure of the training and evaluation data is treated as a central design choice for improving model robustness. Random splits are avoided because they can inflate performance estimates in geospatial applications due to spatial autocorrelation \cite{roberts2017crossvalidation,roscher2024better}. Instead, the data within each study country are partitioned into four non-overlapping regional groups, labeled A-D in Fig.~\ref{fig:socioeconomic_distribution_cmr_ssd_zmb}.

For each fold, the corresponding regional groups from South Sudan, Cameroon, and Zambia are pooled according to their assigned roles, and a cross-country model within each fold is trained. For example, in one fold, Groups~C and D from all three countries (combined) form the training set, Group~A is used for validation, and Group~B is reserved exclusively for testing.

Within each fold, the training, validation, and test regions are spatially disjoint. The validation set is used for model selection, whereas the test set remains excluded from model development and is used only for final evaluation. This design allows the model to learn from broad spatial and demographic variation across the three countries while evaluating its performance on geographically distinct regions not used during training. The partitioning is designed to preserve three forms of representation:

\begin{itemize}
    \item [i)] \textbf{Variation in level of development}, covering urban, peri-urban, and rural areas,    
    \item [ii)] \textbf{Population-group diversity}, including refugees, former refugees, internally displaced persons, refugee returnees, and host communities, and 
    \item [iii)] \textbf{Ecological and climatic heterogeneity} across the study area.
\end{itemize}

The validation strategy is partially blocked validation (spatially blocked but not temporally blocked)\footnote{Partially blocked validation in this context means that spatial dependence is fully blocked by assigning geographically distinct regions to separate training, validation, and test partitions, whereas temporal blocking is only partial because observations from different months and survey years are randomly distributed across these partitions rather than held out as independent survey periods.} \cite{roberts2017crossvalidation}. We address spatial dependence through the regional partitioning described above, while assigning observations collected across different months and survey years to the training, validation, and test sets via a random split rather than temporal blocking. Because the pooled dataset spans fieldwork from 2023 to 2025, including extended survey periods such as the South Sudan FDS from April 2023 to December 2023, this design captures some within- and between-year temporal variation but does not test generalization to an entirely independent future survey round. As additional FDS rounds become available, the framework can be extended to fully blocked spatiotemporal validation across both regions and survey periods.


\subsection{Model architecture}
\label{subsec:model_architecture}

\begin{figure*}[t]
    \centering
    \footnotesize
        \includegraphics[trim=0.2cm 0.01cm 0.1cm 0.1cm,clip]{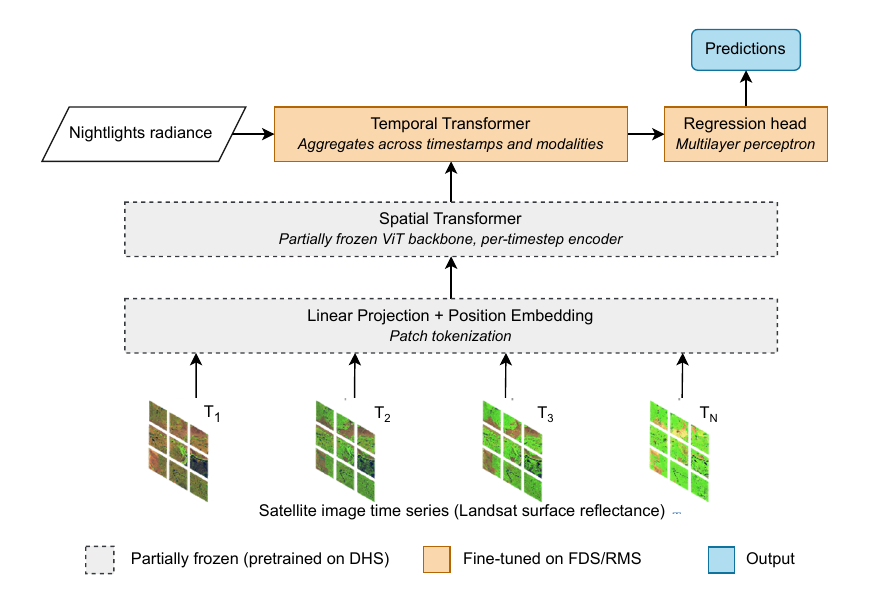} 
    
    \caption{Overview of the multimodal spatiotemporal transformer architecture. Landsat surface-reflectance time series are encoded by a spatial transformer initialized from DHS-based pretraining. A subset of the spatial layers is retained as frozen, while the remaining spatial layers, temporal transformer and regression head are fine-tuned using FDS and RMS data. VIIRS night-time radiance is incorporated as an auxiliary input to the temporal transformer, and building features are added to the temporal transformer output in the regression head.}
    \label{fig:Vits_architecture}
\end{figure*}

We formulate the task as a regression problem in which a sequence of satellite observations and auxiliary geospatial features estimates the grid-level socioeconomic index defined in Section~\ref{subsec:socioeconomic_index_construction}. The model is a multimodal spatiotemporal vision transformer based on the satellite image time-series framework of Pettersson et al. ~\cite{pettersson2023timeseries}, with separate stages for spatial encoding and temporal aggregation, as illustrated in Fig.~\ref{fig:Vits_architecture}.

The primary input is a Landsat surface-reflectance time series, with each observation representing a $6.72~\text{km} \times 6.72~\text{km}$ image tile. We divide each image into patches, project them into latent embeddings, and combine them with positional information following the Vision Transformer formulation \cite{dosovitskiy2020image}. A spatial transformer encodes each acquisition independently, producing one representation per timestamp. A temporal transformer then processes these representations and aggregates information across acquisition dates. VIIRS night-time radiance is incorporated as an auxiliary input to the temporal transformer, providing complementary information on electrification and economic activity. The temporal transformer output is passed to a multilayer perceptron regression head, where building features, representing settlement structure, are added before the socioeconomic index is predicted.

\subsection{Transfer learning and fine-tuning}
\label{subsec:transfer_learning}

Training the spatiotemporal transformer from scratch is impractical given the limited number of grid-level FDS and RMS observations. We therefore initialize the model from the pretrained architecture in Pettersson et al. ~\cite{pettersson2023timeseries}, which was developed using data from approximately 1.2 million households across 140 DHS surveys in 36 African countries between 1990 and 2020. We use a partial fine-tuning strategy to balance knowledge transfer and adaptation to forced displacement settings. The patch-embedding components and early spatial-transformer layers retain their pretrained weights, while the remaining spatial layers, temporal transformer, auxiliary-feature components, and regression head are fine-tuned using the FDS and RMS data. The number of frozen early spatial-transformer layers was treated as a
hyperparameter during model selection. This design preserves general visual representations learned from large-scale DHS imagery while allowing higher-level spatial, temporal, and multimodal relationships to adapt to the distinctive settlement patterns and socioeconomic conditions of forcibly displaced and host populations. Restricting the number of trainable parameters also reduces the risk of overfitting to the comparatively small target dataset.

\subsection{Performance metrics for evaluation}
\label{sec:evaluation_metrics}

We assess the model’s predictive performance using standard regression metrics, including mean absolute error (MAE), mean squared error, root mean squared error, the coefficient of determination ($R^2$), and the Pearson correlation coefficient. Collectively, these metrics provide complementary perspectives on prediction error and goodness-of-fit, enabling evaluation of model performance in predicting the target socioeconomic indicators against ground-truth measurements.

\section{Conclusion}
\label{sec:conclusion}

Earth observation and machine learning have been applied with growing success to poverty estimation in low- and middle-income countries, yet their adaptation to the data-scarce environments of refugee and host community settlements remains largely unexplored. This study addresses that gap. We show that a spatiotemporal vision transformer pretrained on large-scale DHS household data can be adapted, through partial fine-tuning, to predict socioeconomic welfare among forcibly displaced populations across South Sudan, Cameroon, and Zambia. The findings provide proof-of-concept evidence that representations learned from national survey data retain predictive value after adaptation to displacement settings.

The framework offers a scalable pathway to operationalize the growing body of high-quality, publicly available microdata curated by the World Bank\footnote{\url{https://microdata.worldbank.org/index.php/catalog?page=1&sk=refugees&sort_by=rank&sort_order=desc&ps=15}} and UNHCR\footnote{\url{https://microdata.unhcr.org/index.php/home}}  \cite{masaki2023data}. New satellite imagery enables the model to estimate how selected indicators vary across locations and change over time. These spatially granular, regularly updated estimates extend the value of periodic surveys between rounds or when follow-up surveys are not feasible.

Operationally, the framework is most useful as a forced displacement area screening tool to identify where medium-term socioeconomic conditions may be changing and where further assessment or field verification is warranted. Household surveys remain the benchmark for measuring household welfare and are essential for periodically recalibrating the model. As additional FDS and RMS rounds become available, spatiotemporal validation could support applying the framework to displacement contexts beyond sub-Saharan Africa.

\showmatmethods{} 

\dataavail{The FDS and RMS microdata used in this study are available through the UNHCR Microdata Library (\url{https://microdata.unhcr.org/index.php/home}), subject to application, approval, and applicable data-protection requirements.} 

\acknow{This research was supported by the World Bank-UNHCR Joint Data Center on Forced Displacement (JDC). We thank the members of the project's technical advisory Group - David Newhouse, Erwin Knippenberg, Hai-Anh H. Dang, Andrea Pellandra, and Giulia Del Panta, Geraldine Henningsen - for their valuable technical advice on model development, training-data strategy, and validation design. The findings, interpretations, and conclusions expressed in this paper are entirely those of the authors and do not necessarily represent the views of the World Bank Group, the UN Refugee Agency (UNHCR), or their affiliated organizations.\newline
Project title: Transfer Learning for High-Resolution Socio-Economic Data. Project code: P169210}

\showacknow{} 

\newpage
\bibsplit[23]

Bibliography

\bibliography{pnas-sample}

\newpage
\section{Appendix}
\label{sec:Appendix}

 
\begin{table}[!h]
\centering
\footnotesize
\caption{Zero-shot model performance}
\label{tab:zero_shot_model_performance_summary}
\begin{tabular}{p{2.0cm}p{0.7cm} p{0.7cm} p{0.7cm}p{0.7cm}p{1.0cm}}
\toprule
Country & MAE & MSE & RMSE & $R^2$ & Correlation \\
\midrule

South Sudan  & 11.40  &  212.04  &  14.56  &   0.08  &  0.33\\
 
Cameroon & 6.97 & 80.81  & 8.99 &  0.05 & 0.41\\

Zambia & 8.12  &  84.49 & 9.19  &  -0.49  &  0.21 \\
\bottomrule
\end{tabular}
\end{table}

\begin{table}[!h]
\centering
\footnotesize
\caption{Model performance in South Sudan, Cameroon and Zambia on the test data.}
\label{tab:model_performance_refugees_groupB}
\begin{tabular}{p{1.7cm}p{0.8cm} p{0.8cm} p{0.8cm}p{0.8cm}p{1.0cm}}
\toprule
Country & MAE & MSE & RMSE & $R^2$ & Correlation \\
\midrule
South Sudan & 5.99 & 55.23 & 7.43 &	0.26 & 0.52 \\
Cameroon & 5.21 & 41.49	& 6.44 & 0.43 & 0.66 \\
Zambia & 5.42 & 48.05 & 6.93 & 0.41 & 0.65 \\
\bottomrule
\end{tabular}
\end{table}

        

        
        


        
                

\begin{table}[!h]
\centering
\footnotesize
\caption{Socioeconomic index across refugee camps in Mayukwayukwa, Meheba, and Mantapala in  Zambia.}
\label{tab:zambia_camp_socioeconomic_summary}
\begin{tabular}{p{3.0cm}p{1.0cm} p{1.0cm} p{1.0cm}}
    \toprule
    Refugee camp & N & Mean & Std \\
    \midrule
    Mayukwayukwa & 684  & 17.28 & 8.18 \\
    Meheba       & 1025 & 17.55 & 9.52 \\
    Mantapala    & 450  & 15.87 & 6.35 \\
    \bottomrule
\end{tabular}
\end{table}

\begin{figure}[!h]
    \centering
    \footnotesize
    \includegraphics[
            width=\linewidth,
            keepaspectratio]{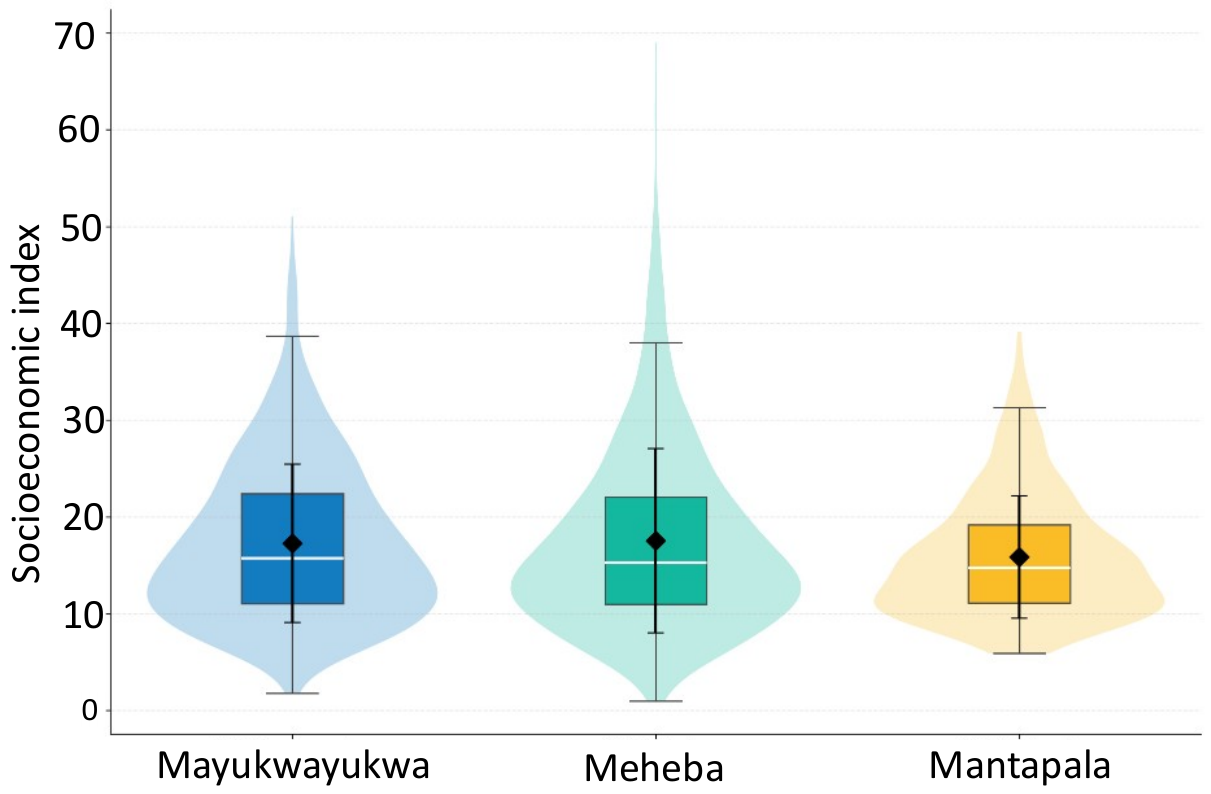}
        \captionof{figure}{\footnotesize
        Socioeconomic index distribution across refugee camps in Mayukwayukwa, Meheba, and Mantapala in  Zambia.
        }
        \label{fig:socioeconomic_distribution_zambia_camps}
   
\end{figure}

\begin{figure}[ht]
    \centering
    \footnotesize
    \vspace{-17cm}
    \includegraphics[scale=0.40]{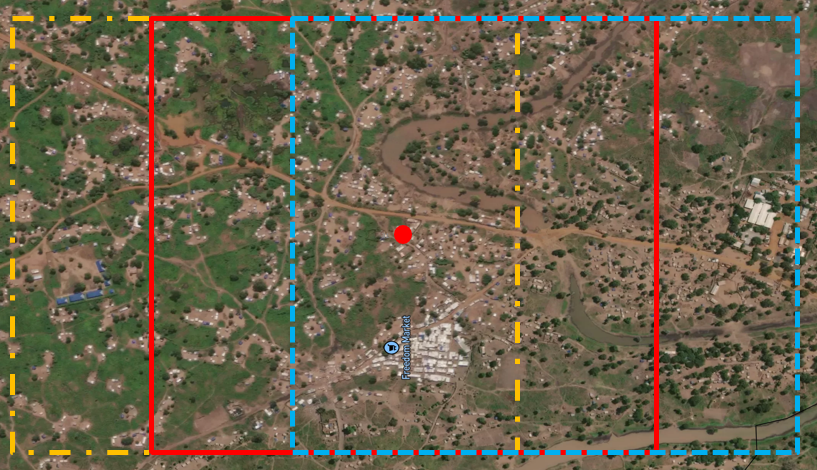}
     \caption{Illustration of the sliding-window augmentation strategy. The solid red grid represents the original extraction window, and the yellow dash-dotted and blue dashed grids represent spatially shifted windows. The red dot marks the center of the original grid.}
\label{fig:FDS_data_aug}
\end{figure}

\end{document}